\documentclass[conference]{IEEEtran}
\IEEEoverridecommandlockouts
\usepackage{cite}
\usepackage{amsmath,amssymb,amsfonts}
\usepackage{algorithmic}
\usepackage{graphicx}
\usepackage{textcomp}
\usepackage{xcolor}
\usepackage{hyperref}
\usepackage{caption}
\usepackage{subcaption}

\DeclareMathOperator*{\argmax}{arg\,max}
\DeclareMathOperator*{\argmin}{arg\,min}

\def\BibTeX{{\rm B\kern-.05em{\sc i\kern-.025em b}\kern-.08em
    T\kern-.1667em\lower.7ex\hbox{E}\kern-.125emX}}
    
\begin{document}

\title{Preference-conditioned Pixel-based AI Agent For Game Testing}

\author{
\IEEEauthorblockN{Sherif Abdelfattah\textsuperscript{\textsection}}
\IEEEauthorblockA{\textit{Xbox Studios Quality AI Lab} \\
sherifgad@microsoft.com}
\and
\IEEEauthorblockN{Adrian Brown}
\IEEEauthorblockA{\textit{Xbox Studios Quality AI Lab} \\
adrianbr@microsoft.com}
\and
\IEEEauthorblockN{Pushi Zhang}
\IEEEauthorblockA{\textit{Microsoft Research Asia} \\
pushizhang@microsoft.com}
}

\IEEEoverridecommandlockouts

\IEEEpubid{\makebox[\columnwidth]{000-0-0000-0000-0/00/$00.00$~\copyright2023 IEEE \hfill} 
\hspace{\columnsep}\makebox[\columnwidth]{ }}

\maketitle
\IEEEpubidadjcol
\begingroup\renewcommand\thefootnote{\textsection}
\footnotetext{Corresponding author}
\endgroup

\begin{abstract}
The game industry is challenged to cope with increasing growth in demand and game complexity while maintaining acceptable quality standards for released games. Classic approaches solely depending on human efforts for quality assurance and game testing do not scale effectively in terms of time and cost. Game-testing AI agents that learn by interaction with the environment have the potential to mitigate these challenges with good scalability properties on time and costs. However, most recent work in this direction depends on game state information for the agent's state representation, which limits generalization across different game scenarios. Moreover, game test engineers usually prefer exploring a game in a specific style, such as exploring the golden path, yet, current game testing AI agents do not provide an explicit way to satisfy such a preference. This paper addresses these limitations by proposing an agent design that mainly depends on pixel-based state observations while exploring the environment conditioned on a user's preference specified by demonstration trajectories. In addition, we propose an imitation learning method that couples self-supervised and supervised learning objectives to enhance the quality of imitation behaviors. Our agent significantly outperforms state-of-the-art pixel-based game testing agents over exploration coverage and test execution quality when evaluated on a complex open-world environment resembling many aspects of real AAA games.
\end{abstract}


\section{Introduction}
\label{sec:introduction}
The gaming industry has grown significantly over the last decade, with expected revenue to exceed \$$320$ billion by $2026$~\cite{WorldEco_games22}. In addition, there has been a steady increase in the player base with an accompanying rise in the 
 complexity of games. As a result, game developers have struggled to meet this increasing demand while preserving the quality of the delivered games. Conventional game testing activities involving humans are costly and inflexible to scale with the expanding market and game complexity trends~\cite{game_test_survey21}. Thus, there is a need to automate such game testing activities in a cost-efficient and scalable manner.

Artificial intelligence (AI) agents that can adapt based on interaction with the game environment can mitigate the game quality assurance challenges in a scalable and cost-effective manner. Reinforcement learning (RL)~\cite{SuttonB98} provides an optimization framework to train such agents interactively guided by a reward signal representing the quality of the taken actions w.r.t the task's objective. RL agents have achieved many breakthroughs in playing video games with human-level performance~\cite{rl_1_13,rl_2_5,rl_3_21,rl_4_21,rl_5_20}. This has recently motivated attempts to adopt RL agents in the game testing domain~\cite{CCPT_2022,game_test_aug_2020,game_test_coverage2021,game_test_reveal2019,game_test_survey21,game_test_wuji}. Despite the demonstrated effectiveness of these methods, they usually depend on game's internal information (e.g., semantic maps or 3D positioning)  to formulate the agent's state representation. This dependency limits the agent's ability to generalize across different scenarios that do not share the same internal information. In addition, these methods do not support game object testing for the lack of a formal way to condition the test on specific objects.

A recent RL agent called \emph{Inspector}~\cite{inspector22} addressed these limitations by depending only on pixel-based information (i.e., RGB frames) to formulate the agent's state. Moreover, the Inspector agent proposed a practical framework combining novelty-seeking exploration, imitation learning for test execution in a human-like style, and object detection for identifying objects to test. While showing promising results, the Inspector agent follows visual novelty for exploration without a way to condition the exploration behavior on a given preference, such as exploring across a game's golden path (i.e., standard playable path). This is vital for adapting to fluctuating test requirements and achieving sample-efficient exploration in open-world games. Moreover, we notice that most RL game testing methods lack the ability to focus the visual state representation learning on relevant test aspects in the environment, which is essential for high-quality test imitation and exploration.

This paper proposes a Preference-conditioned Pixel-based Game Testing Agent (PPGTA) that mitigates the mentioned limitations in existing game testing RL agents by extending the Inspector agent's framework. We summarize our contribution points as follows:

\begin{itemize}
    \item We propose a novel technique to condition the agent's exploration behavior on a preferred style.
    \item We present an effective recurrent architecture for imitation learning that counts for global and local context information within a trajectory.
    \item We show a novel way to incorporate a self-supervised consistency regularization objective during imitation learning to maximize training signals and focus the visual feature extraction on test-relevant aspects.
    \item We propose a design for estimating novelty rewards based on ensemble agreement for better robustness against the non-stationary nature of novelty reward functions.
\end{itemize}

The remainder of this paper is organized as follows. Section~\ref{sec:related_work} reviews the related work. Section~\ref{sec:problem_definition} formulates the research problem. Section~\ref{sec:methodology} introduces our methodology. Section~\ref{sec:experimental} describes the experimental design. Section~\ref{sec:results_disc} presents the evaluation results and discusses the findings. Finally, Section~\ref{sec:conclusion} concludes the work and highlights future directions to extend it.

\section{Related Work}
\label{sec:related_work}
Automated game testing is a challenging task, usually involving  maximizing game space coverage and performing testing. Over the last period, many AI-based methods were proposed to tackle this task. Chang et al.~\cite{game_test_reveal2019} proposed a game exploration method called \emph{Reveal-More} that augments human tester efforts to maximize game coverage. Their approach depends on an initial exploration executed manually by a human tester to identify viable states for exploration trials; then, a Rapidly-Exploring Random Trees (RRT) algorithm~\cite{LaValleK99} is used to perform random exploration walks starting for the set of viable states. Zheng et al.~\cite{game_test_wuji} proposed an evolutionary optimization method representing game testing as a multi-objective problem with exploration and gameplay objectives. The authors couple RL and evolutionary selection to evolve a set of non-dominated policies that maximize both objectives over an iterative procedure.

Bergdahl et al.~\cite{game_test_aug_2020} utilized deep RL algorithms to run multiple goal-seeking agents simultaneously to exploit unexpected paths in a given game using distance from the goal point as a reward function. Gordillo et al.~\cite{game_test_coverage2021} proposed a curiosity-driven RL agent that maximizes game test coverage by running multiple agents simultaneously guided by a count-based curiosity reward calculated on the discretized 3D positions. Visitation graphs were used to identify anomalous trajectories. Tufano et al.~\cite{game_load_ai_22} proposed an RL agent targeting game load testing to identify performance issues. The authors combined game score and load rewards to promote an effective adversarial behavior that maximizes gameplay while exploiting performance issues.

Despite the effectiveness of the previous RL methods, they depend on the game's internal state (e.g., global positioning, semantic maps, game scores) to represent the state, which limits their ability to generalize across different games. Recently, a work called \emph{Inspector}~\cite{inspector22} proposed a novel framework for training pixel-based game testing RL agents. It jointly trains three components: a curiosity-driven exploration policy, an object detection model to identify objects of interest for testing, and an imitation policy to execute test behaviors. The agent showed a promising performance on object-centric game testing. Yet, the Inspector agent does not have an explicit way to condition the exploration policy on a given preference to control its behavior. Moreover, it used a simple novelty estimation model~\cite{RND19} prone to catastrophic forgetting in open-world environments~\cite{GoExplore21}. We note a recent work~\cite{CCPT_2022} that proposed a way to condition the exploration policy on a given style, yet it is not pixel-based and not suitable for object testing. Our PPGTA agent addresses these limitations by extending the Inspector framework to accept a given exploration style through demonstrations and use a more robust novelty estimator while working with pixel-based input.

\section{Problem Definition}
\label{sec:problem_definition}
Our problem definition comes in two folds: learning to explore the environment and learning to imitate expert test demonstrations. We aim to explore the environment guided by an exploration reward signal. We formulate this as a standard Markov Decision Process (MDP) setup $<S,A,P,R,\mu,\gamma>$, where $S$ is a state space, $A$ is an action space, $P(s_t+1|a_t,s_t)$ is a state transition probability distribution, $r_t=R(s_t,a_t)$ is a reward function, $s_0\sim\mu$ is the initial state probability distribution, and $\gamma$ is a reward discounting factor. We target optimizing the parameters $\omega$ of an exploration policy $a_t=\pi^e_{\omega}(s_t)$ by maximizing a discounted reward return:

\begin{equation}
    \label{eq:rl_obj}
    \argmax_{\omega^*\in\Omega}\, \sum_{t=1}^T \gamma{R(s_t,\pi^e_{\omega}(s_t))}
\end{equation}

\noindent where $T$ is the time horizon.

While for the imitating expert demonstrations, we follow a supervised learning setup. Given an expert demonstration dataset $D=\{\tau^1,\tau^2,\dots,\tau^N\}$ where $\tau^i=\{(a^p_1,s_1),(a^p_1,s_1),\dots,(a^p_K,s_K)\}$ is a trajectory of state-action pairs with a length $K$, we target optimizing the parameters $\theta$ of an imitation policy $\hat{a^p}_t=\pi^i_{\theta}(s_t)$ by minimizing a cross-entropy loss $H$ with the expert's policy $a^p_t=\pi^p(s_t)$:

\begin{equation}
    \label{eq:imi_loss}
    \argmin_{\theta^*\in\theta}\, \sum_{n=1}^N\sum_{(a^p_t,s_t)\in\tau^n} H(\pi^i_{\theta}(s_t),a^p_t)
\end{equation}

\section{Methodology}
\label{sec:methodology}

\begin{figure*}[htbp]
\centerline{\includegraphics[scale=0.359]{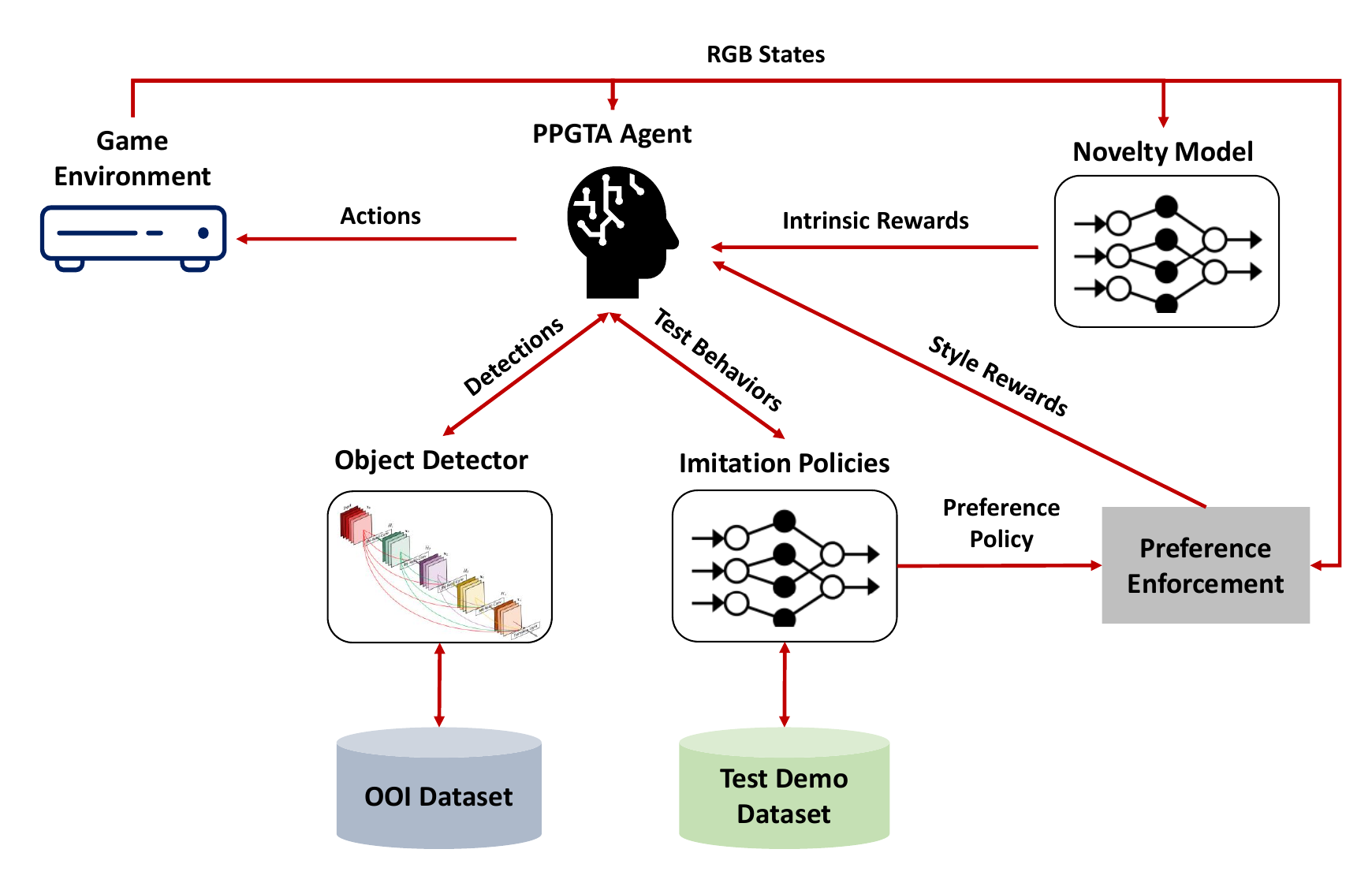}}
\caption{A high-level block diagram of the Preference-conditioned Pixel-based Game Testing Agent (PPGTA). The novelty model rewards exploration based on scene novelty. The imitation policy executes bug testing. The object detection model detects objects of interest (OOIs) to trigger bug testing. The preference enforcement module conditions the exploration behavior.}
\label{fig:main_framework}
\end{figure*}

This section introduces details of our Preference-conditioned Pixel-based Game Testing Agent (PPGTA) following the \emph{Inspector} framework~\cite{inspector22}, including three main building blocks. First, we describe our technique for learning object-aware visual features by fine-tuning a pre-trained image feature extractor following an inverse dynamics prediction objective. Second, we present the details of training our imitation learning policy that selects actions in an auto-regressive manner using Gated Recurrent Units (GRU)~\cite{GRU14} memory cells and regularized by a temporal object-aware consistency regularization objective. Finally, we introduce an effective way to condition the environment exploration policy on a user's preference presented in the form of a few expert demonstrations.

\subsection{Feature Distillation Via Masked Inverse Dynamics Learning}
\label{sec:method_FE}
Learning to distill relevant visual features is a vital objective in reinforcement learning for robustness against variations in visual dynamics (e.g., background aspects, lighting, etc.) and learning in a  sample-efficient manner. A common way to achieve this visual distillation is following an inverse-dynamics objective~\cite{PathakAED17} by learning to predict actions conditioned on their adjacent states. This helps to focus the feature encoder (FE) on controllable aspects of the state. 

\begin{figure}[htbp]
\centerline{\includegraphics[scale=0.3]{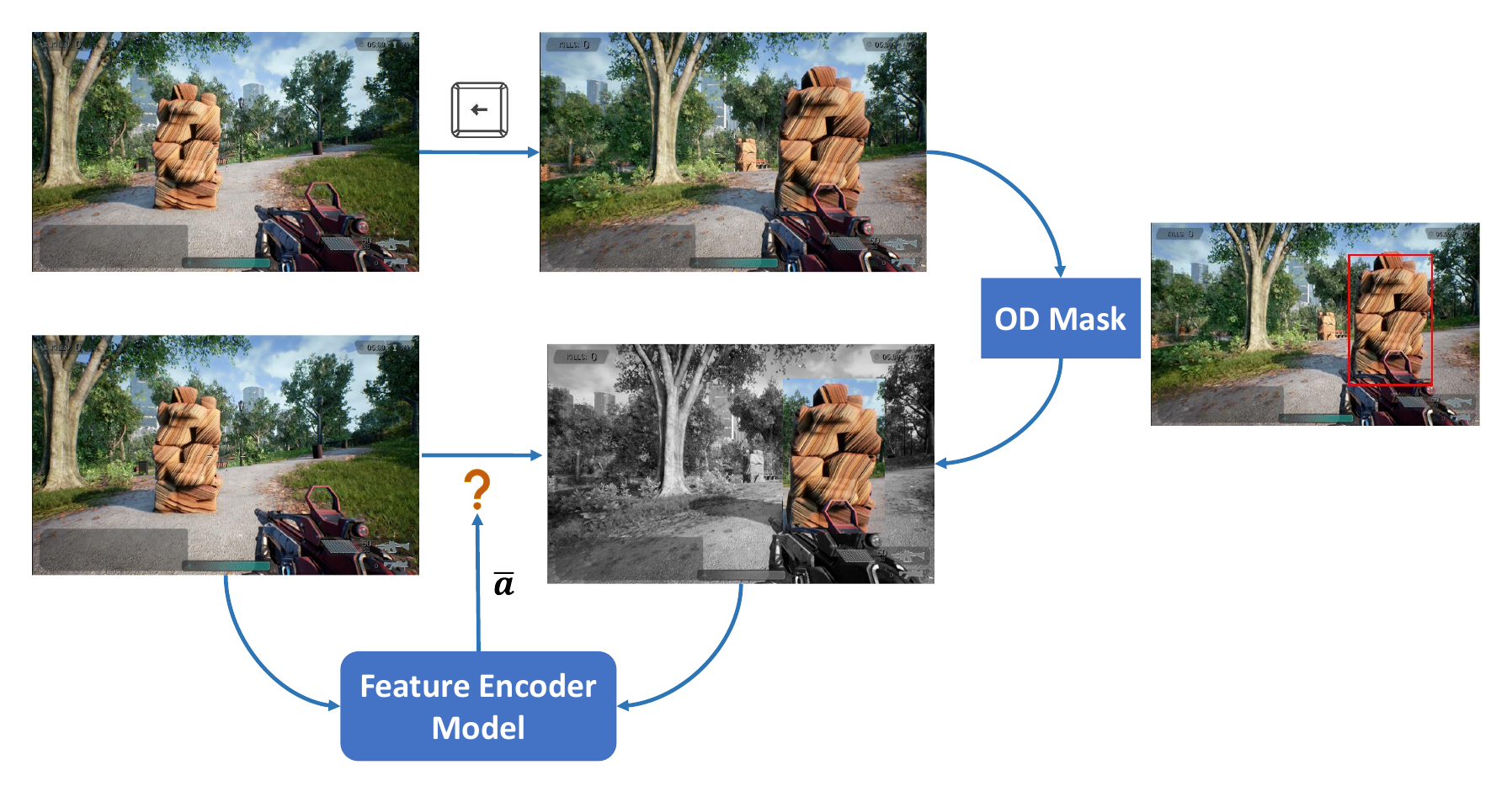}}
\caption{Our visual encoder model training using masked inverse dynamics prediction objective. Best viewed in colors.}
\label{fig:fe_model}
\end{figure}

Our setup assumes an object bug-testing scenario over objects of interest (OOIs) in the environment. Following a similar approach to the Inspector framework~\cite{inspector22}, we use an object detection (OD) model to identify OOIs while exploring the environment. We adopt a few-shot tuning approach~\cite{FsDET20} that mixes samples from pre-training datasets and a downstream dataset to prevent catastrophic forgetting and speed up the fine-tuning procedure. We utilize a Faster R-CNN model~\cite{faster_rcnn_17} pre-trained on the COCO dataset~\cite{COCO14}. Leveraging the existence of an OD model, we create object-aware augmentations for RGB frames by masking them around detected OOIs. Precisely, our OD masking technique converts a colored image into grayscale except for the region identified by the bounding box of the OOI under test (see Figure~\ref{fig:fe_model}). As in Figure~\ref{fig:fe_model}, we randomly augment one of the two adjacent frames during the inverse-dynamics training using our OD mask and fine-tune a pre-trained visual feature encoder by minimizing an action prediction loss as per Equation~\ref{eq:fe_loss}.

\begin{equation}
    \label{eq:fe_loss}
    \argmin_{{\theta^*}\in\Theta} \mathcal{L}(a_t,\phi_\theta(s_{t-1},s_t))
\end{equation}

\noindent where $\mathcal{L}$ is a categorical cross-entropy loss function, $a_t$ is the action taken at time $t$, $\phi_\theta$ is a visual encoder model with $\theta$ parameters, $\Theta$ is the parameter search space, and $(s_{t-1},s_t)$ is a tuple of two adjacent states before and after taking action $a_t$.

\subsection{Memory-augmented Imitation Learning}
\label{sec:method_IL}
\begin{figure}[htbp]
\centerline{\includegraphics[scale=0.37]{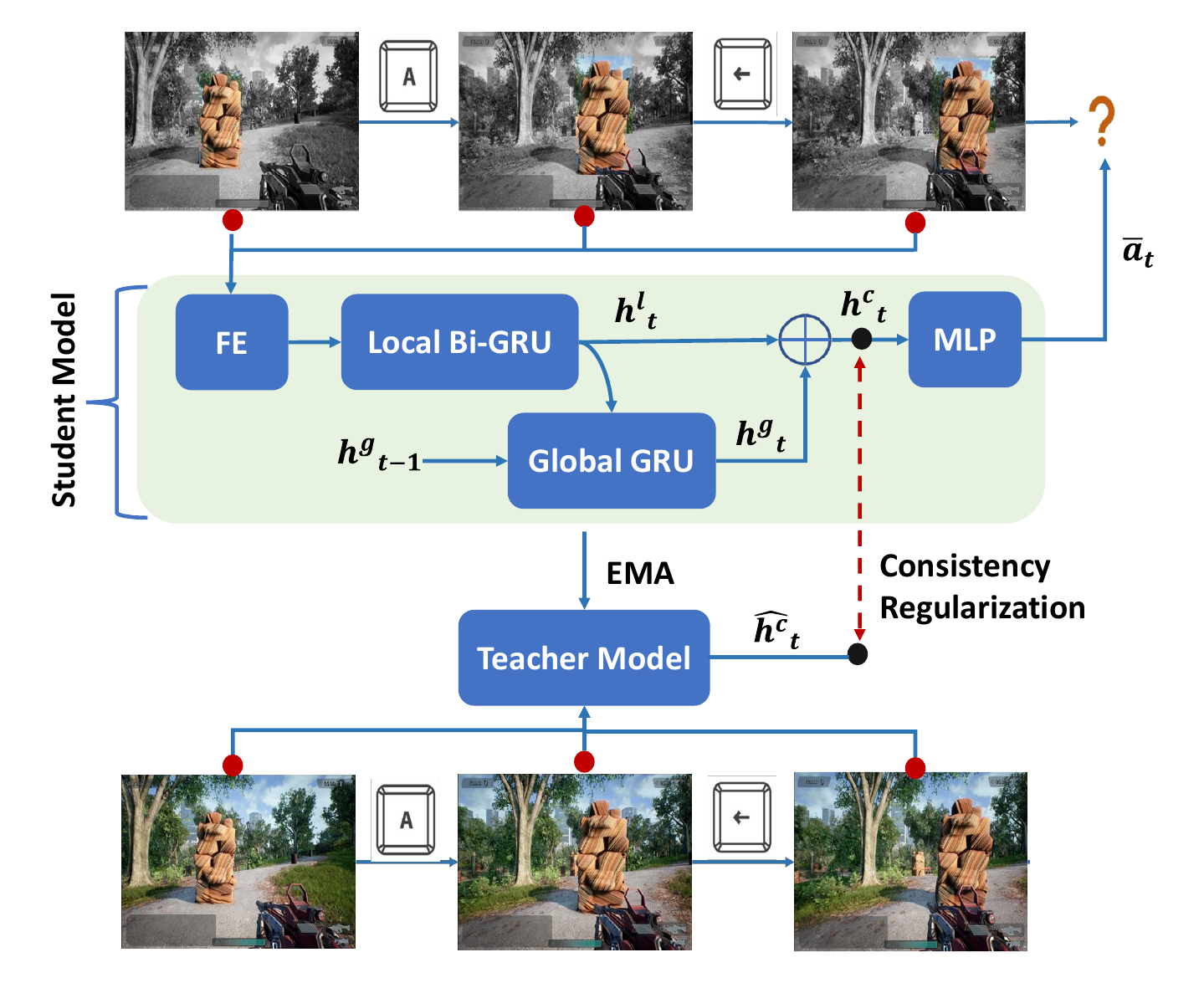}}
\caption{Training procedure for our memory-augmented imitation learning model. Best viewed in colors.}
\label{fig:il_model}
\end{figure}

We build on the fine-tuned visual feature encoder to design an imitation learning policy that learns to mimic human game testing from demonstrations. To count for object-aware temporal dynamics in a test trajectory sample, we extend the DINO~\cite{DINO21} self-supervised objective to work in a temporal setup. Figure~\ref{fig:il_model} illustrates our imitation learning training setup, where a student model consists of four building blocks, including:
\begin{itemize}
    \item A visual feature encoder (FE) encodes the last $K$ adjacent frames in a training trajectory masked using the previously described OD mask technique.
    \item A local bi-directional gated recurrent unit (GRU) cell that encodes temporal dynamics across the last $K$ frames into a state $h^l$.
    \item A global GRU cell that encodes temporal dynamics across the whole trajectory till the current time point into state $h^g$.
    \item An action prediction feed-forward head (MLP) that gets a concatenated state $h^c$ from both local and global GRU cells as input.
\end{itemize}

Similar to DINO~\cite{DINO21}, our teacher model $f^t_\omega$ is a version of the student model $f^s_\omega$ updated using an exponential moving average (EMA) estimate. Moreover, the teacher model gets full RGB frames as input instead of masked ones compared to the student model. We use consistency regularization between the concatenated states across student and teacher models as an auxiliary objective to enforce object-aware temporal dynamics modeling. In this stage, we minimize the following loss function.

\begin{equation}
    \label{eq:il_loss}
    \argmin_{{\omega_s^*}\in\Omega} \mathcal{L}(a_t,f^s_\omega(h^c_t)) - \lambda{(\hat{h^c}_t\log{h^c_t})}
\end{equation}

\noindent where $\mathcal{L}$ is a categorical cross-entropy loss function, $a_t$ is the action taken at time $t$, $f^s_\omega$ is the student model parameterized with $\omega$, $h^c$ is the student's concatenated memory state, $\lambda$ is a regularization parameter, and $\hat{h^c}_t$ is the teacher's concatenated memory state.

\subsection{Preference-Conditioned Environment Exploration}
\label{sec:method_explore}

\begin{figure}[htbp]
\centerline{\includegraphics[scale=0.35]{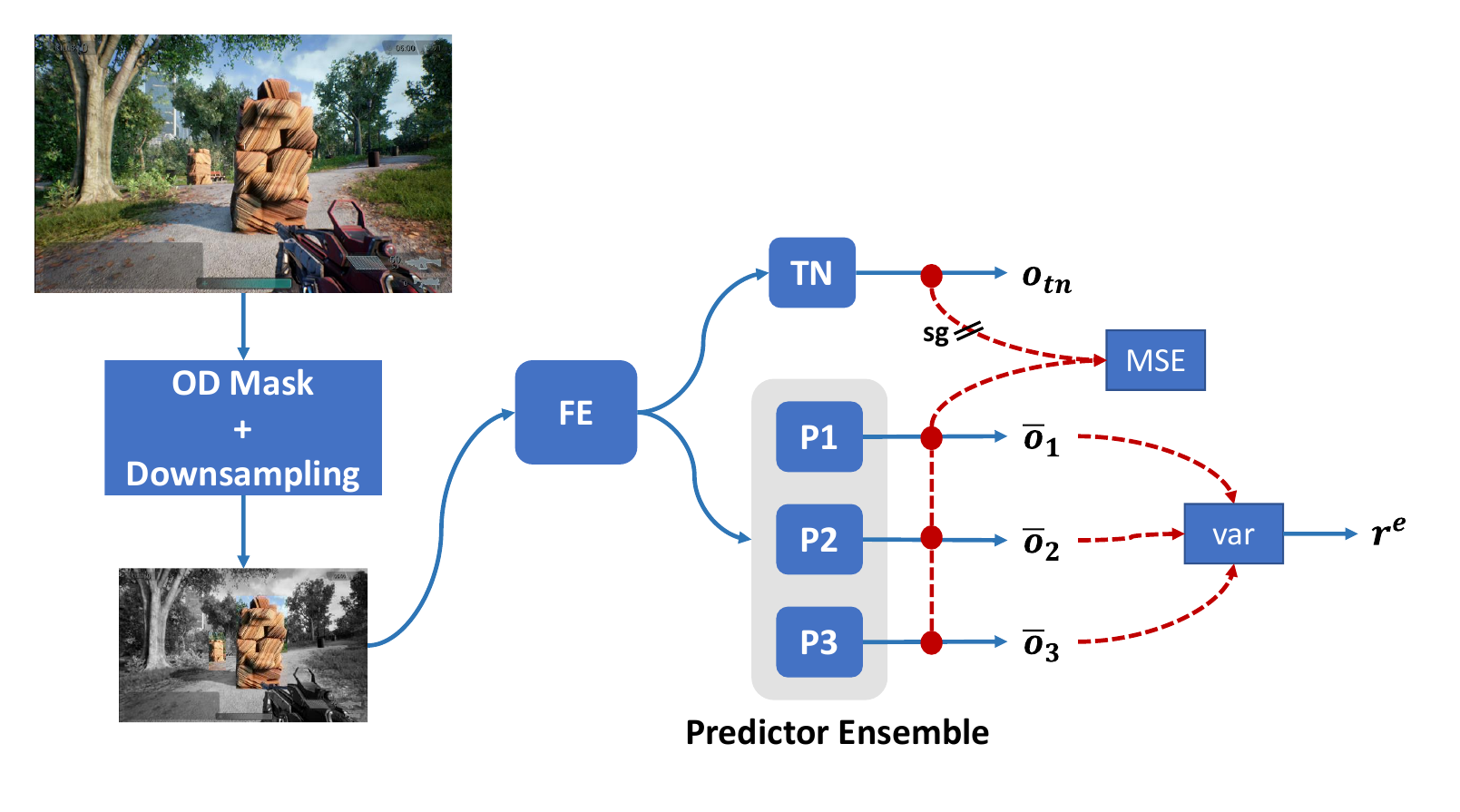}}
\caption{Our extended RND model for generating exploration intrinsic rewards. Best viewed in colors.}
\label{fig:RND_model}
\end{figure}

Our exploration policy follows an intrinsic state novelty reward signal that promotes visiting unexplored regions in the environment. Random Network Distillation (RND)~\cite{RND19} is a commonly used method to generate this reward. The RND workflow involves two networks, including a randomly initialized and fixed target network $TN$ that takes an input state and outputs a projection for it $o_{tn}$ and a predictor network $P$ that outputs a prediction for the target projection $\hat{o}$ given the input state. The reward is directly proportional to the predictor network's error. RND could suffer from catastrophic forgetting~\cite{NGU20} in complex and open-world environments. To overcome this limitation, we extend it in two steps, including 1) downsampling the input RGB image and masking it using our OD masking method to reduce variance and facilitate a better novelty  prediction~\cite{GoExplore21}, and 2) utilizing an ensemble of predictors $\{P_1,P_2,\dots,P_N\}$ instead of a single one as in the original RND design and using the prediction variance across ensemble to generate the intrinsic reward $r^e_t = Var(\hat{o}_1,\hat{o}_2,\dots,\hat{o}_N)$. We optimize the parameters of the predictor ensemble using the Mean Squared Error (MSE) loss between target and predicted projections $MSE(o_{tn},\hat{o})$. Figure~\ref{fig:RND_model} illustrates our extended RND.

Directly following an intrinsic reward signal does not necessarily result in a favorable exploration behavior from a test engineer's perspective. To facilitate a preference-conditioned exploration behavior, we condition the exploration policy $\pi^e$ on a user's preference using an exploration style reward $r^p$. In our experimental setup, we define the preference of following the pathway using a path-following imitation learning policy $\pi^p$ and calculate the style reward $r^p=-KL(\pi^e_{1:t},\pi^p_{1:t})$ using Kullback–Leibler (KL) divergence between action distributions of the exploration policy $\pi^e$ and path-following policy $\pi^p$ till time point $t$. 

We use a linear combination of the intrinsic reward $r^e$ and the style reward $r^p$ as the final reward $r^c=\alpha{r^p}+(1-\alpha)r^e$, where $\alpha\in[0,1]$ is a mixing hyper-parameter. We initialize $\alpha$ with the value of $0.8$ and design an adaptive technique to adjust its value during exploration based on the change in novelty reward $r^e$ over steps. In the case of decreasing novelty of the space (e.g., hitting an invisible wall), we decrease $\alpha$ with a rate of $0.05$ until it reaches a minimum threshold of $0.5$ to facilitate a more exploratory behavior. We quantify a decrease in $r^e$ by measuring the change over adjacent transitions $r^e_t-r^e_{t-1}<\epsilon$ for $\epsilon=-0.1$. Once the change in $r^e$ rises, we increase $\alpha$ with a rate of $0.1$ until it reaches the max value of $0.8$. Our exploration policy $\pi^e$ is trained to maximize the discounted combined reward return $\sum_{t=1}^T\,\gamma{r^c_t}$ using the PPO algorithm~\cite{PPO17}.

\section{Experimental Design}
\label{sec:experimental}
In this section, we introduce the details of our experimental evaluation design. We describe the open-world environment setup and OOI distribution. Then, we elaborate on the training datasets for object detection and imitation learning. Afterward, we inform on the utilized performance metrics and the comparative baseline. Finally, we provide implementation details for our PPGTA agent.

\subsection{Environment Setup}
\begin{figure}[htbp]
\centerline{\includegraphics[scale=0.58]{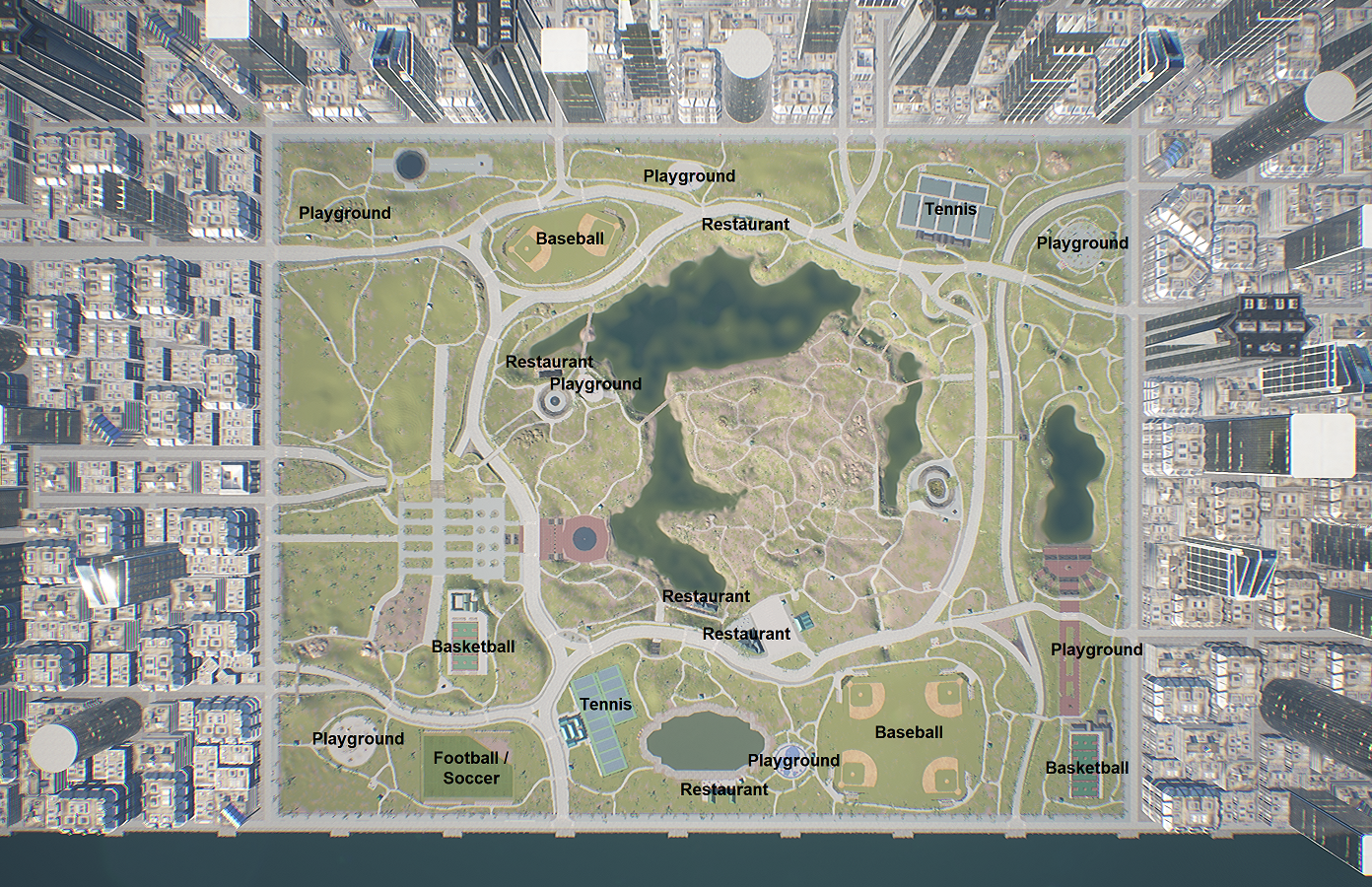}}
\caption{A birds-eye view of our experimental open-world environment resembling a large city park region.}
\label{fig:exp_env}
\end{figure}

To effectively evaluate the performance of the PPGTA agent, we design an open-world environment using the \emph{Unreal} engine~\footnote{\url{https://www.unrealengine.com/en-US/}} as one of the most utilized game development engines. Our environment resembles the characteristics of a big city park that spans over $3.3$ square kilometers (km2) and includes pathways, green spaces, lakes, service areas, and bridges. Figure~\ref{fig:exp_env} shows a birds-eye view of the environment. The environment has four OOI types, including rock pillars, barrels, tires, and chairs. We randomly distribute them across pathways in the park. We introduced two Object bug types, including low-resolution and stretched texture bugs. We create a $50\%$-$50\%$ split between normal and bugged instances for each OOI type in the object distribution. 

\subsection{Training Datasets}
For fine-tuning our OOI detection model, we collect a small-sized and balanced dataset of $2000$ shots for the four OOI types in our environment. We split this dataset using a stratified sampling of $70\%$-$10\%$-$20\%$ for training, validation, and test sets respectively.

\begin{figure*}[htbp]
\centerline{\includegraphics[width=\textwidth, height=80px]{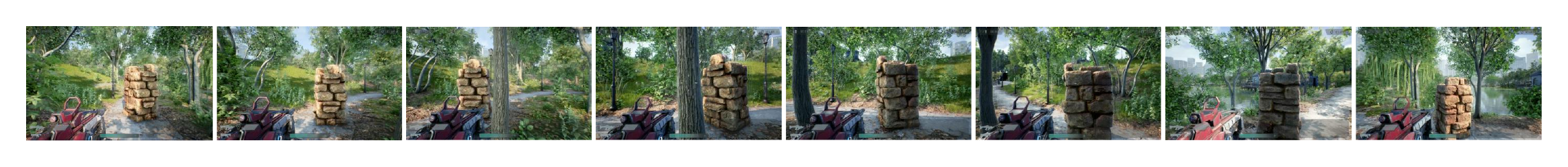}}
\caption{Screenshots from a sample 360 degrees texture bug testing behavior executed on a rock pillar object by the PPGTA agent. Best viewed in colors.}
\label{fig:texture_test}
\end{figure*}

We collect two datasets for imitation learning training, one for texture bug testing behavior and another for path-following behavior as an exploration preference. The texture bug testing behavior is executed by moving around an OOI $360$ degrees to cover all its faces (see Figure~\ref{fig:texture_test}), and we collect a total of $150$ trajectories (i.e., sequences of RGB frame and action pairs) performed by test engineers. The path-following behavior is performed as following the pathways in a limited region of the environment, We note that this region is discarded during agent evaluation to prevent it from using memorized training scenes as an advantage. We collect $50$ path following trajectories, each with a length of $2$ minutes. We combine RGB frames and actions from the two imitation datasets to build a dataset for the inverse-dynamics training of the feature extractor (FE) backbone model. Upon training, the FE model is used as the backbone within the imitation policy network during imitation learning training. We follow a $70\%$-$10\%$-$20\%$ train-validation-test splitting for imitation learning and inverse-dynamics datasets. 

\subsection{Performance Evaluation Setup}
We design an open-world game testing task that involves navigating the environment to find OOIs and performing texture bug testing on them. The character in the game is a first-person shooter spawn adopted from the Unreal engine Shooter demo~\footnote{\url{https://www.unrealengine.com/marketplace/en-US/product/shooter-game}}. We define a discrete action space of $9$ actions using keyboard keys, including the WASD keys for moving in four directions, the Arrow keys for camera rotation in four directions, and the Space Bar key for jumping. For the open-ended nature of this task, we limit the time horizon for each episode to $2000$ steps. We execute $45$ episodes, each with a different starting position in the environment. We uniformly randomize starting positions across the environment to promote diverse exploration.

We compare our PPGTA agent with the Inspector agent~\cite{inspector22} as a recent state-of-the-art method for object bug testing in AAA games. Besides, this comparison highlights our contributions in extending Inspector's limitations. We evaluate the performance of exploration and bug testing objectives using the percentage of explored space and the number of tested OOIs metrics. We visualize these metrics using birds-eye maps from the evaluation environment.

\subsection{Implementation Details}

For the object detection fine-tuning, we use a mini-batch size of $128$, a learning rate of $3e^{-4}$, an input RGB frame size of $128\times128\times3$, and $4$ epochs. While for imitation learning, we use a stack of $4$ RGB frames of size $128\times128\times3$ as input, a learning rate with a linear warm-up from $2e^{-3}$ to $1e^{-2}$ during the first $10$ epochs, and a cosine decay thereafter, a mini-batch size of $64$, and a total of $40$ epochs with early stopping. We empirically selected AlexNet~\cite{AlexNet12} pre-trained model as our feature extractor (FE) backbone as it achieved the best performance and efficiency tradeoff compared with well-known pre-trained models, including RestNet-18, ResNet-50, and EfficientNet. As in Figure\ref{fig:il_model}, we follow the FE backbone with two bi-directional GRU cells for local and global views with hidden state sizes of $256$ and $1024$ respectively followed by two multilayer-perceptron (MLP) layers with sizes $512$ and $256$ respectively. We use layer-norm on the two MLP layers and a dropout of $0.2$. 

For the extended RND model, we use the FE backbone followed by two multilayer-perceptron (MLP) layers with sizes $512$ and $256$ respectively, for both the target and predictor network architectures. The RND training data come from a replay buffer of size $1024$ transitions, and its input frames are downsized into $64\times64\times3$ and masked using the OD mask. For the PPO actor-critic networks, we follow the same architecture of the imitation learning policy with two different heads for action and value predictions. We also initialize actor and critic networks using the parameter values of the path-following imitation learning policy as an informative inductive bias. We use the same PPO hyper-parameter setup as the Inspector's PPO agent~\cite{inspector22} and follow its original configuration for the hyperparameters of the Inspector baseline. We use the \emph{AdamW} algorithm~\cite{AdamW19} for all optimization objectives. We utilize \emph{UnrealCV} plugin~\footnote{\url{https://unrealcv.org/}} to integrate with the game environment.

\section{Results and Discussion}
\label{sec:results_disc}
This section presents the experimental setup results and discusses the findings. We discuss the findings across three evaluations: exploration performance evaluation, bug testing imitation performance evaluation, and object-detection evaluation. 

\subsection{Exploration Performance Analysis}
\begin{figure}
     \centering
     \begin{subfigure}[b]{0.45\textwidth}
         \centering
         \vspace{-13mm}
         \includegraphics[width=0.9\textwidth, height=183px]{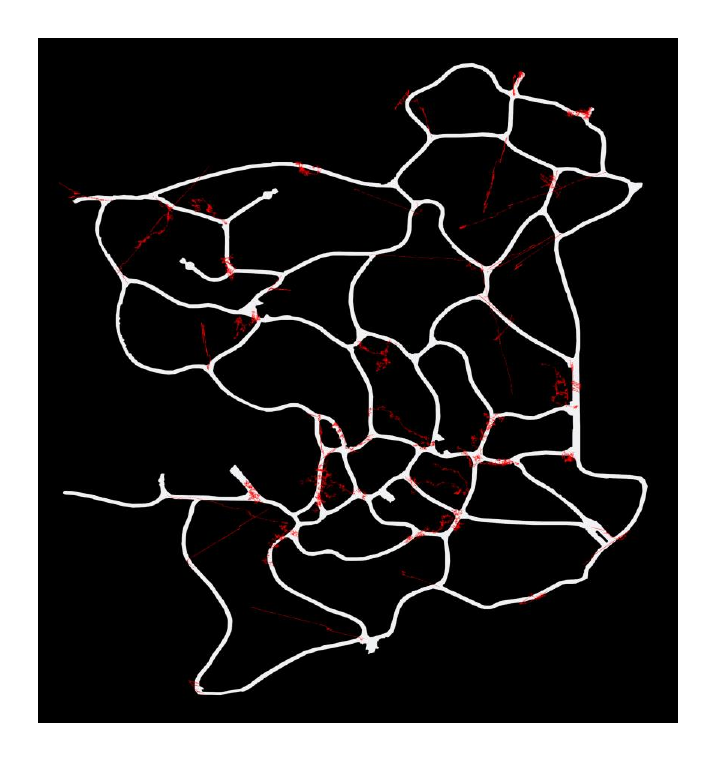}
         \caption{}
         \label{fig:inspector_exploration}
     \end{subfigure}
     \begin{subfigure}[b]{0.45\textwidth}
         \centering
         \includegraphics[width=0.9\textwidth, height=183px]{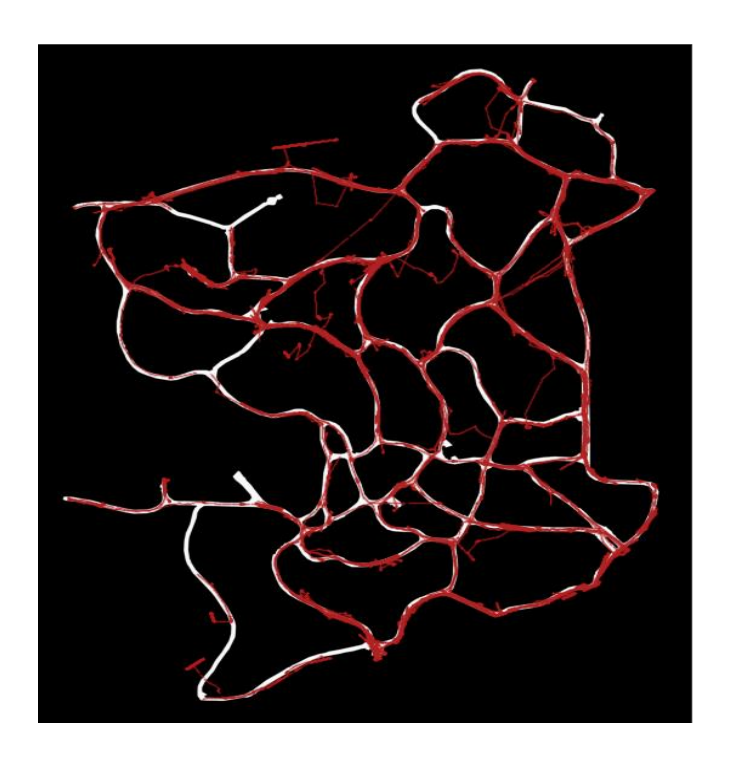}
         \caption{}
         \label{fig:ppgta_exploration}
     \end{subfigure}
\caption{Two birds-eye view maps highlighting pathways in the experimental environment in white lines and agent's exploration traces in red lines. (a) Exploration traces for the Inspector agent. (b) Exploration traces for the PPGTA agent. Best viewed in colors.}
        \label{fig:exploration_evaluation}
\end{figure}

We compare the exploration performance between the PPGTA and Inspector agents over $45$ episodes; each spans $2000$ steps and starts from a different position. Figure~\ref{fig:exploration_evaluation} depicts two birds-eye view maps showing exploration traces in red lines over the pathways in the environment for the Inspector agent (Figure~\ref{fig:inspector_exploration}) and the PPGTA agent (Figure~\ref{fig:ppgta_exploration}). Our PPGTA agent significantly outperforms the Inspector agent in terms of the preferred area (i.e., pathways) coverage after $45$ episodes. We quantified the covered pathways area to be $80\%$ in the PPGTA compared to $16\%$ in the Inspector agent. This observation highlights the benefits of having a preference-conditioned exploration policy via a balance between style and novelty rewards compared to depending solely on novelty rewards in open-world environments.

\begin{figure}[htbp]
\centerline{\includegraphics[scale=0.65]{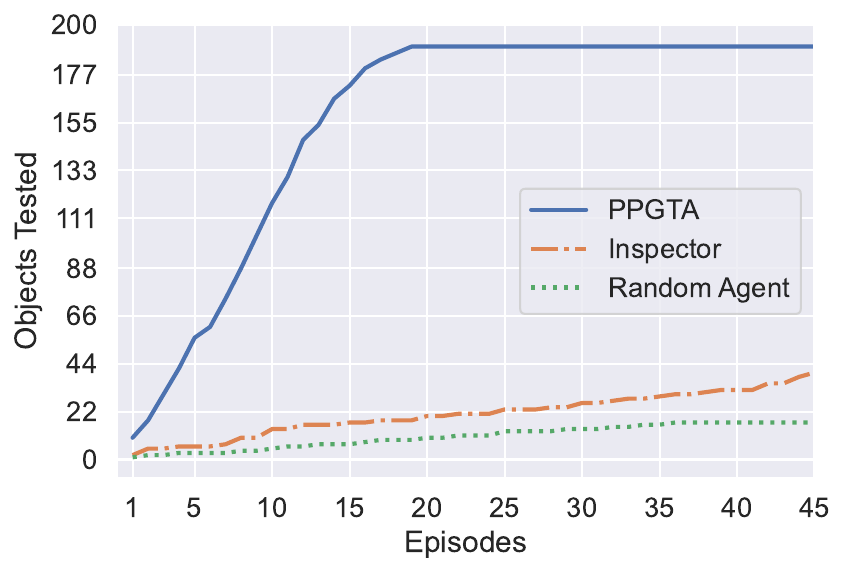}}
\caption{A line plot showing the number of tested OOIs over the $45$ exploration episodes for each agent.}
\label{fig:exp_ooi_test_rate}
\end{figure}

We assess the OOI testing rate over the $45$ episodes of exploration for PPGTA, Inspector, and a random agent baseline to further highlight the sample efficiency of PPGTA. Figure~\ref{fig:exp_ooi_test_rate} shows a line plot for the number of tested OOIs over episodes. We observe that PPGTA took $18$ episodes to test $192$ OOIs out of a total of $247$ OOIs in the environment compared to $19$ OOIs and $9$ OOIs tested by the Inspector and random agents respectively at this stage. After $45$ episodes, the total OOIs covered by the Inspector and random agents were $40$ and $17$ respectively, which is significantly lower than the number of tested OOIs by the PPGTA agent. Another interesting finding was observed after investigating the remaining OOIs missed by the PPGTA agent, as we found the majority of these objects was unreachable due to invisible walls or pathway blockages by other objects. This was a valid navigation bug in the object distribution.

\subsection{Imitation Learning Performance Analysis}

To evaluate the quality of performed bug testing behaviors, we design a practical way to measure the similarity between the expert's and agent's actions distributions. We use \emph{Jensen-Shannon} divergence~\cite{Nielsen19} between the action distribution from an executed test behavior and the action distribution from all expert demonstrations of that behavior. We empirically identified a success threshold on the \emph{Jensen-Shannon} divergence value based on the average value achieved by the imitation policy of each agent on validation trajectories after the last training epoch.

\begin{figure}
     \centering
     \begin{subfigure}[b]{0.45\textwidth}
         \centering
         \includegraphics[width=0.9\textwidth, height=183px]{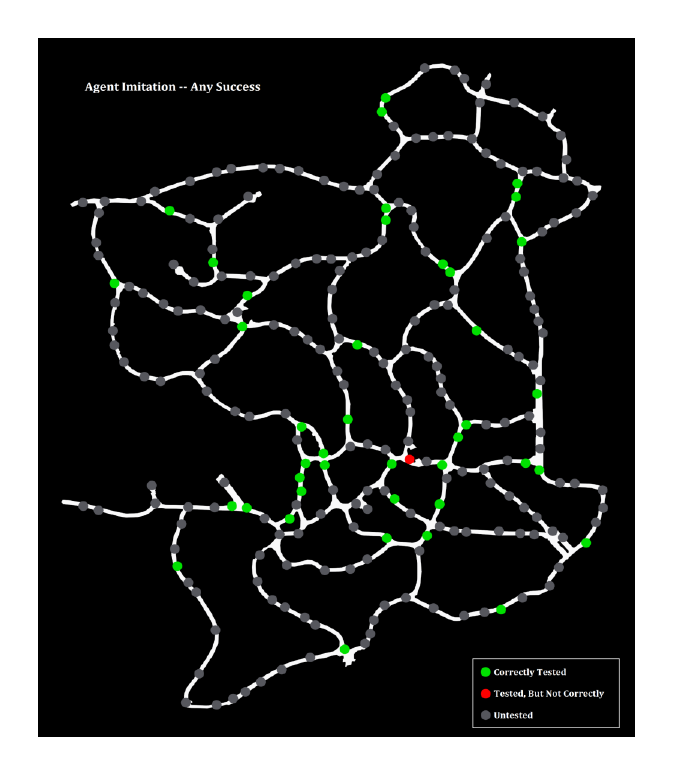}
         \caption{}
         \label{fig:inspector_imitation}
     \end{subfigure}
     \begin{subfigure}[b]{0.45\textwidth}
         \centering
         \includegraphics[width=0.9\textwidth, height=183px]{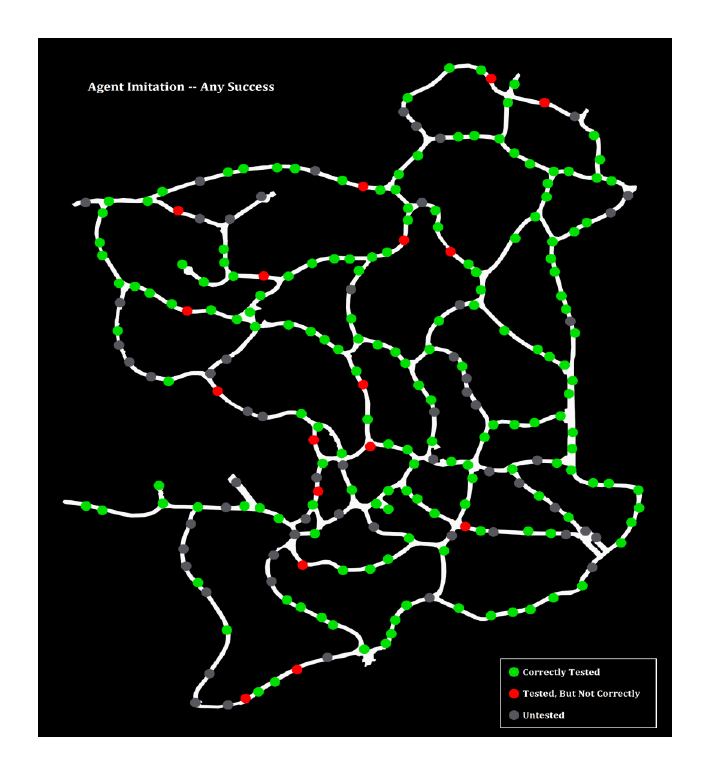}
         \caption{}
         \label{fig:ppgta_imitation}
     \end{subfigure}
\caption{Two birds-eye view maps highlighting the status of bug testing across object distribution in the environment, including success tests in green color, failed tests in red color and missed tests in gray color. (a) The Inspector agent's tests. (b) The PPGTA agent's tests. Best viewed in colors.}
        \label{fig:imitation_evaluations}
\end{figure}

We plot the object distribution of the experimental environment setup using birds-eye maps of the pathways represented as dots and use color coding to discriminate between successful bug tests with green color, failed bug tests with red color, and missed bug tests with gray color. Figure~\ref{fig:imitation_evaluations} shows two maps showing the status of bug testing execution attempts comparing the Inspector agent (Figure~\ref{fig:inspector_imitation}) and the PPGTA agent (Figure~\ref{fig:ppgta_imitation}). The PPGTA agent achieved a success rate (i.e., the ratio of correctly tested OOIs over the total OOIs) of $71\%$ compared to a $16\%$ success rate by the Inspector agent. This finding indicates a significant increase in the test success rate of $55\%$. This is a result of the enhanced performance of both exploration and imitation policies of the PPGTA agent compared to Inspector's ones.

\subsection{OOI Detection Analysis}

We evaluate the performance of the OOI detection model using the average precision (AP) metric~\cite{AP20} across different bounding box sizes in the test object detection dataset. Table~\ref{tbl:ap_results} summarizes the AP results across the four OOI types in our experimental design. While fine-tuning on a low data regime with around 500 samples for each OOI type, the detection model achieved acceptable AP results across all types that are sufficient to perform the task effectively. This finding supports the ability of the OOI detection model to generalize across an arbitrary number of types while minimizing data labeling costs.

\begin{table}[htbp]
\caption{Test average precision (AP) results for OOI detection.}
\label{tbl:ap_results}
\begin{center}
\vspace{-5mm}
\begin{tabular}{|c|c|c|c|c|}
\hline
\textbf{}&\multicolumn{4}{|c|}{\textbf{OOI Types}} \\
\cline{2-5} 
\textbf{Metric} & \textbf{\textit{Rock Pillar}}& \textbf{\textit{Barrel}}& \textbf{\textit{Tire}} & \textbf{\textit{Chair}}\\
\hline
AP& $74.5\%$& $73.8\%$ & $71.8\%$ & $72.6\%$\\
\hline
\end{tabular}
\label{tab1}
\end{center}
\end{table}
\vspace{-5px}

\section{Conclusion}
\label{sec:conclusion}

In this paper, we proposed a sample efficient yet effective game testing agent called \emph{PPGTA}. Our agent addressed limitations in existing game testing agents through three contribution points. First, it is mainly working with pixel-based observations reducing the burden on the game developer to provide internal game state observations. Second, it proposes an easy way to condition its exploration behavior according to the test scenario needs by providing a small number of demonstrations representing the required exploration style. Third, it presents a novel recurrent policy architecture that learns from the local and global contexts in the trajectory while following a temporal regularization objective to focus its visual feature extractor on important state aspects. Fourth, it presents a robust design for estimating intrinsic exploration rewards using agreement over an ensemble of observation predictors. We evaluated our PPGTA agent on an open-world game environment with an object bug testing task and compared it to a state-of-the-art pixel-based game testing agent and a random baseline. Results showed that our agent significantly outperforms other comparatives on exploration coverage and imitation test quality while being sample-efficient.

In future work, we aim to explore learning a generic skill set based on expert demonstration datasets; thus, we can target imitating complex test behaviors across multiple tasks and games. Also, we will explore model-based exploration by training world models simultaneously during imitation learning. Finally, we will consider more flexible alternatives to generalize on an arbitrary number of objects of interest (OOIs), such as conditioning on language prompts.

\bibliographystyle{IEEEtran}
\bibliography{main}

\end{document}